\documentclass[conference]{IEEEtran}
\IEEEoverridecommandlockouts

\usepackage{cite}
\usepackage{amsmath,amssymb,amsfonts}
\usepackage{algorithmic}
\usepackage{graphicx}
\usepackage{textcomp}
\usepackage{xcolor}
\usepackage{booktabs}
\usepackage{multirow}
\usepackage{colortbl}
\usepackage{url}
\usepackage[most]{tcolorbox}
\usepackage{tabularx}
\usepackage{float}

\def\BibTeX{{\rm B\kern-.05em{\sc i\kern-.025em b}\kern-.08em
    T\kern-.1667em\lower.7ex\hbox{E}\kern-.125emX}}
\begin{document}

\title{Abstract Compression of Conversational Audio Context for LLM-based ASR
}
\author{
\IEEEauthorblockN{
Shashi Kumar$^{1,2}$,
Esaú Villatoro-Tello$^{1}$,
Sergio Burdisso$^{1}$,
Kadri Hacioglu$^{3}$,
Thibault Bañeras-Roux$^{1}$,\\
Hasindri Watawana$^{1,2}$,
Dairazalia Sanchez-Cortes$^{1}$,
Srikanth Madikeri$^{4}$,
Petr Motlicek$^{1,5}$,
Andreas Stolcke$^{3}$
}
\vspace{3mm}
\IEEEauthorblockA{
$^{1}$\textit{Idiap Research Institute, Switzerland};
$^{2}$\textit{EPFL, Switzerland};
$^{3}$\textit{Uniphore, U.S.A.}\\
$^{4}$\textit{University of Zurich, Switzerland};
$^{5}$\textit{Brno University of Technology, Czech Republic}
}
}

\maketitle

\begin{abstract}
Standard LLM-based speech recognition systems process utterances in isolation, limiting their ability to leverage conversational context. Here we study whether multimodal context from prior turns improves LLM-based ASR and how to represent that context efficiently. We find that, after supervised multi-turn training, conversational context mainly helps with the recognition of contextual entities. However, conditioning on raw context is expensive because the prior-turn audio embedding sequence grows rapidly with conversation length. To address this, we propose Abstract Compression, a learned latent abstraction of prior-turn audio that replaces each prior-turn audio sequence with a fixed number of latent tokens while retaining corresponding transcripts explicitly.
On both in-domain and out-of-domain test sets, the compression-enabled model recovers part of the gains of raw-context conditioning with a smaller prior-turn audio footprint. We also provide targeted analyses of the compression setup and its tradeoffs.
\end{abstract}

\begin{IEEEkeywords}
Context-aware ASR, speech recognition, multimodal LLMs, audio compression, long-context modeling.
\end{IEEEkeywords}

\section{Introduction}
\label{sec:intro}
Automatic speech recognition (ASR) is increasingly deployed in conversational settings, such as voice assistants, customer-support calls, meetings, and multimodal agents. In these scenarios, interpreting an utterance often depends on previous turns, which can establish domain-specific terminology, introduce named entities, or provide discourse cues that resolve ambiguity. Despite this dependence, most ASR systems process utterances independently \cite{kim2018dialog, kim2019gated, hori2021advanced, lee2024enhancing}. This limitation is especially detrimental for rare, error-prone contextual entities (e.g., names, locations, products). Prior work on contextual and conversational ASR has addressed this through text-side contextual biasing via finite-state transducers, shallow fusion, or context-aware end-to-end architectures \cite{pundak2018deep, williams2018contextual, jain2020contextual, tong2023hierarchical, zhou2024copyne, li2024cb, liu2024post}, as well as session-level language modeling \cite{xiong2018session}. While effective, these methods focus on text-side biasing signals or single-modality history rather than rich, multimodal conversational context.
 
\begin{figure*}[t!]
    \centering
    \includegraphics[width=\linewidth]{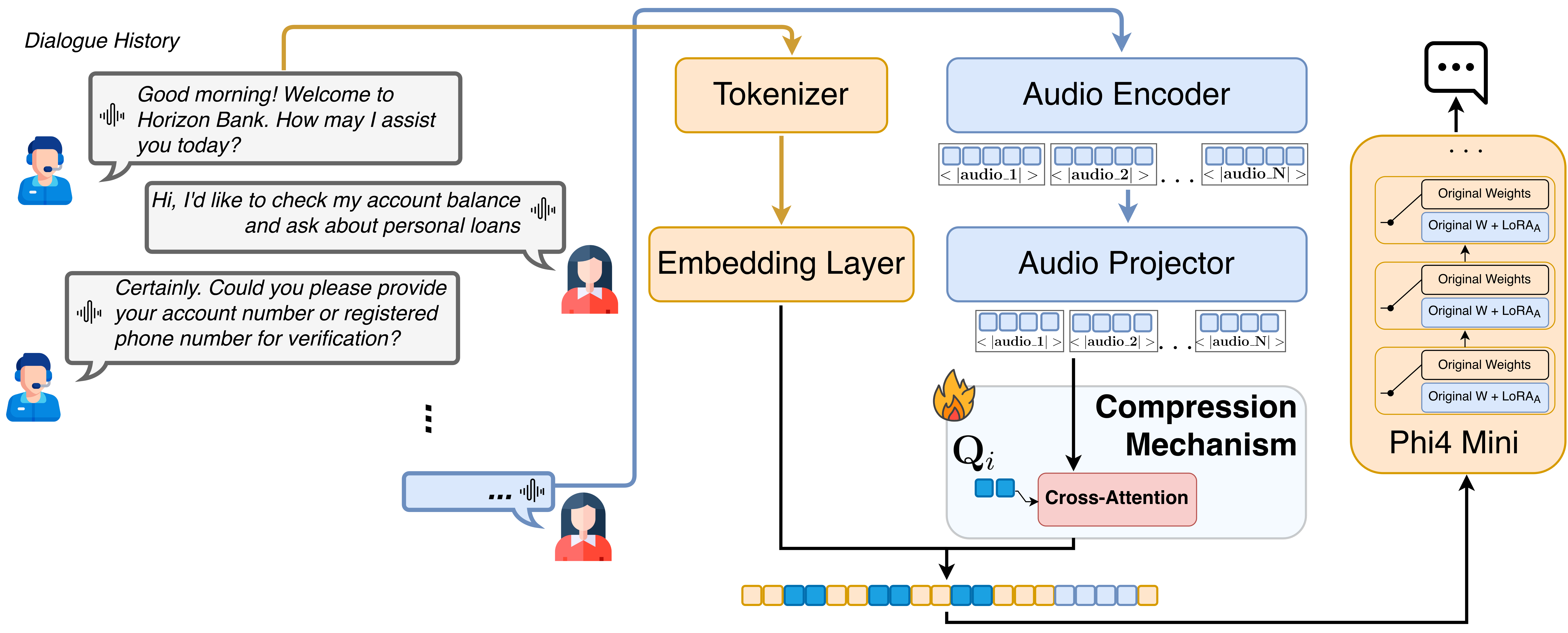}
\caption{\textbf{Overview of Abstract Compression for context-aware ASR.}
Prior conversational turns are represented by both transcript and audio. In our implementation, the audio from each prior turn is distilled into a fixed number of latent tokens, while transcript is retained explicitly. These compressed context representations are provided to the LLM alongside the current turn’s full-resolution audio embeddings. This preserves part of the conversational context while reducing the cost of raw-context conditioning.}
    \label{fig:method_overview}
\end{figure*}

Multimodal large language models (LLMs) provide a natural framework for revisiting this problem \cite{tangsalmonn, wu2023decoder, ma2024embarrassingly, kumar2024performance, abouelenin2025phi, burdisso2025_text_only, lakomkin2024end, villatoro_2026_context_proj}. By mapping audio into the embedding space of an autoregressive LLM, these systems can natively process interleaved audio, text, and structured prompts. Most prior work in LLM-based ASR has focused on single-utterance recognition or general instruction-following. Extending this to conversational ASR allows prior turns to provide both lexical evidence (what was said via word choice) and acoustic evidence (how it was said via audio). However, exploiting this multimodal context introduces a severe computational bottleneck: prior-turn audio is represented by long sequences of tokens, causing prompt lengths, key-value (KV) cache costs, and inference latency to grow rapidly with conversation length. 

While general long-context modeling and learned compression, such as cross-attention bottlenecks \cite{jaegle2021perceiver, alayrac2022flamingo, li2023blip2} and memory summarization \cite{Rae2020Compressive, xu2023recomp, li2025atacompressor, lin2025refragrethinkingragbased}, have been explored for text and vision, how to effectively distill multimodal context specifically for ASR remains an open question. This raises two linked questions: Does conversational multimodal context actually improve LLM-based ASR? And can those gains be retained under a fixed, substantially smaller context budget?

In this work, we study this tradeoff. We first demonstrate that simply prepending raw conversational context degrades performance at inference time. However, after supervised multi-turn fine-tuning, the model benefits significantly from context, showing distinct gains on recovering contextual entities (measured via bias-word error rate). Because raw-audio context conditioning is expensive, we introduce \textbf{Abstract Compression}, a task-driven bottleneck that abstracts each prior-turn audio sequence into a small, fixed number of learned latent tokens while retaining prior-turn transcripts in explicit textual form (see Figure~\ref{fig:method_overview}).

Taken together, our work lies at the intersection of contextual ASR, multimodal LLM-based speech recognition, and efficient sequence modeling. Our specific contributions are:
\begin{itemize}
    \item We show that multimodal context from prior turns improves LLM-based ASR, with the largest gains on contextual entities after supervised multi-turn training.
    \item We propose Abstract Compression to distill prior-turn audio into a compact token footprint, mitigating the primary source of context cost.
    \item Through evaluations on both in-domain and out-of-domain datasets, we show that Abstract Compression recovers a meaningful fraction of the raw-context performance gains, and we provide detailed ablations on the factors governing this quality-efficiency tradeoff.
\end{itemize}

\section{Multimodal LLM-Based ASR}
\label{sec:llm_asr}
In this work, we adopt \textsc{Phi-4-Multimodal} \cite{abouelenin2025phi} as our foundational backbone. Although the architecture supports interleaved image, audio, and text, we focus exclusively on its speech-processing capabilities to cleanly isolate the impact of conversational context on speech recognition performance.

\noindent \textbf{Model Architecture and Notation.} \quad
The model processes an input audio waveform $\mathbf{A}$ through a dedicated audio encoder $f_{\text{enc}}$ and a projection module $f_{\text{proj}}$. For a given audio segment, acoustic features are extracted and projected into the LLM's input embedding space:
\begin{equation}
    \mathbf{X} = f_{\text{proj}}(f_{\text{enc}}(\mathbf{A})),
\end{equation}
where $\mathbf{X} = \{\mathbf{x_1, x_2, \dots, x_L}\}$ represents a sequence of $L$ audio embeddings of dimension $D$. These embeddings are interleaved with text embeddings and processed by the \textsc{Phi-4-Mini} LLM \cite{abouelenin2025phi}. 

\noindent \textbf{Single-Turn ASR Baseline.} \quad
In a standard, context-independent (single-turn) ASR setting, the input sequence $\mathcal{S}_{\text{single}}$ is structured using the model's standard chat template:
\begin{equation}
\label{eq:prompt_asr}
    \mathcal{S}_{\text{single}} = \langle|\text{user}|\rangle \mathbf{X} \backslash n \mathcal{P} \langle|\text{end}|\rangle \langle|\text{assistant}|\rangle,
\end{equation}
where $\mathcal{P}$ is a natural language instruction to transcribe the audio. For all experiments, we use a fixed prompt $\mathcal{P}$: \textit{``Transcribe the audio clip into text.''}, following the \textsc{Phi-4-Multimodal} paper \cite{abouelenin2025phi}.
The text generated by the model immediately following the $\langle|\text{assistant}|\rangle$ token forms our predicted transcription. This single-turn formulation serves as the baseline for all subsequent contextual experiments.

\section{Context-Aware ASR}
\label{sec:context_inc}
To move beyond isolated utterance recognition, we investigate the model's ability to leverage cues from preceding conversational turns. Formally, a conversation is represented as a sequence of $N$ turns, where each turn $i$ consists of an audio segment $\mathbf{A}_i$ and its corresponding transcript $Y_i$. Our objective is to transcribe the $N$-th turn using the preceding $N-1$ turns as context, together with the current audio segment $\mathbf{A}_N$.
Prior turns are indexed relative to the current turn being transcribed, rather than by their absolute positions in the conversation.

\subsection{Inference-Time Context Conditioning}
We first evaluate whether the base model can utilize conversational context through inference-time prompting alone.
For this setting, we follow the model's multi-turn chat prompt format, in which completed prior turns are prepended to the current transcription request.
Let $\mathcal{T}_i$ denote a completed prior turn $i$:
\begin{equation}
\label{eq:single_turn_prompt}
    \mathcal{T}_i = \underbrace{\langle|\text{user}|\rangle \mathbf{X}_i \backslash n \mathcal{P} \langle|\text{end}|\rangle}_{\text{Audio Context}} \underbrace{\langle|\text{assistant}|\rangle Y_i \langle|\text{end}|\rangle}_{\text{Transcription Context}}.
\end{equation}
The full input sequence for the $N$-th turn is the concatenation:
\begin{equation}
\label{eq:multi_turn_prompt}
\begin{split}
\mathcal{S}_{\text{context}} = &\, \mathcal{T}_1 \mathcal{T}_2 \dots \mathcal{T}_{N-1} \\
&\, \langle|\text{user}|\rangle \mathbf{X}_N\backslash n\mathcal{P}
\langle|\text{end}|\rangle \langle|\text{assistant}|\rangle .
\end{split}
\end{equation}
In our experiments, providing the model with raw conversational context at inference time degraded performance relative to the single-turn ASR baseline.
This suggests that inference-time prompting alone is insufficient for reliable cross-turn context use without explicit training.

\subsection{Supervised Fine-Tuning for Contextual Awareness}
We therefore perform supervised fine-tuning (SFT) using the multi-turn format in Eq.~\ref{eq:multi_turn_prompt}. Concretely, the model is trained to predict the transcript of the final turn, $Y_N$, conditioned on the preceding conversational turns $\mathcal{T}_{1}, \dots, \mathcal{T}_{N-1}$ and the current audio input $\mathbf{A}_N$. This allows the model to learn from inputs that include prior turns, rather than from the single-turn formulation alone.

In our experiments, supervised multi-turn fine-tuning with these raw-contexts reduced bias-word error and enabled the model to better recover contextual entities from preceding turns. Detailed results are provided in Section~\ref{subsec:raw_history_results}. However, these gains come at the expense of a substantially larger context size: because each prior-turn audio input $\mathbf{X}_i$ is represented by a high-resolution sequence of audio embeddings, the total length of $\mathcal{S}_{\text{context}}$ grows rapidly with the number of prior turns.

\section{Abstract Compression for Context-Aware ASR}
\label{sec:compression}
In our setting, audio embeddings dominate the prompt length, while transcripts are comparatively compact. Our compression strategy is therefore  to replace only the audio portion of each context turn with a fixed-size latent representation and keep the transcripts explicit.
This focus is complementary to textual context compression via projection \cite{context-projector-IS27}: we target the audio-token bottleneck, while joint use of compressed audio context and projected textual context is left for future work.
Our training strategy is also inspired by recent work on context compression for RAG-based LLM systems \cite{lin2025refragrethinkingragbased}.

\subsection{Compression Mechanism}
Instead of processing the full audio-token sequence of each context turn, we learn a compression function $g(\cdot)$ that maps a variable-length audio sequence to a fixed number of latent tokens.

For each context turn $i$, we compress the high-resolution audio embeddings $\mathbf{X}_i$ into $K$ latent tokens.
In this work, we implement $g(\cdot)$ as a learnable cross-attention mechanism with turn-index-specific query matrices \cite{jaegle2021perceiver}.
These query matrices are defined for relative context positions with respect to the current turn, up to the maximum supported context length.
Concretely, for each turn index $i$, we define a learnable query matrix $\mathbf{Q}_i \in \mathbb{R}^{K \times D}$.
The compressed representation $\mathbf{Z}_i$ is then obtained by cross-attending $\mathbf{Q}_i$ to the original audio embeddings of that turn:
\begin{align}
\mathbf{Z}_{i}
&= \operatorname{CrossAttention}(\mathbf{Q}_{i}, \mathbf{X}_i, \mathbf{X}_i).
\end{align}
Here, $\mathbf{Q}_{i}$ are learnable parameters, but because they attend to turn-specific keys and values ($\mathbf{X}_i$), the resulting $\mathbf{Z}_{i}$ vectors dynamically capture the unique context of that specific turn.
This architecture introduces an information bottleneck that compresses the audio of each prior turn into a fixed-length set of $K$ latent tokens.

\noindent \textbf{Compressed Multi-turn Prompting.} \quad
With the compression module in place, each context turn $\mathcal{T}_i$ in Eq.~\ref{eq:single_turn_prompt} is replaced by its compressed counterpart $\mathcal{T}'_i$:
\begin{equation}
\mathcal{T}'_i
= \langle|\text{user}|\rangle \mathbf{Z}_{i}\backslash n\mathcal{P}
\langle|\text{end}|\rangle\langle|\text{assistant}|\rangle Y_i
\langle|\text{end}|\rangle .
\end{equation}
That is, the original audio-token sequence $\mathbf{X}_i$ is replaced by the compressed representation $\mathbf{Z}_i$, while the transcript $Y_i$ is kept explicit. The resulting compressed multi-turn input sequence is
\begin{equation}
\label{eq:multi_turn_prompt_compressed}
\begin{split}
\mathcal{S}_{\text{comp}} = &\, \mathcal{T}'_1 \mathcal{T}'_2 \dots \mathcal{T}'_{N-1} \\
&\, \langle|\text{user}|\rangle \mathbf{X}_N\backslash n\mathcal{P}
\langle|\text{end}|\rangle \langle|\text{assistant}|\rangle .
\end{split}
\end{equation}
Compression is applied only to prior context turns; the current-turn audio $\mathbf{X}_N$ remains uncompressed to preserve full acoustic detail.

\subsection{Training Strategy}
\label{subsec:compression_train}
Training the model directly on compressed multi-turn inputs is challenging because it must both interpret the latent audio tokens $\mathbf{Z}_i$ and use them for current-turn transcription.
We therefore adopt a two-stage training strategy following \cite{lin2025refragrethinkingragbased}.
In their setting, Stage~1 trains a compression module to reconstruct text from compressed text representations, and Stage~2 performs curriculum-based training with progressively increasing context length.
We adapt this procedure to ASR: Stage~1 aligns compressed audio representations to the LLM through single-turn ASR, and Stage~2 fine-tunes the model on compressed multi-turn inputs using a curriculum over the number of context turns.

\noindent \textbf{Stage~1: Compressed-Audio-to-LLM Alignment.} \quad
The goal of the first stage is to make the compressed audio representations compatible with the LLM input space.
To do so, we repurpose the single-turn ASR task by replacing the raw audio embeddings $\mathbf{X}_1$ in the baseline prompt (Eq.~\ref{eq:prompt_asr}) with the compressed tokens $\mathbf{Z}_1$.
During this stage, the base model is frozen, and only the compression module, including the turn-specific queries $\mathbf{Q}_i$ and the cross-attention parameters, is optimized.
The model is trained with the standard cross-entropy loss to predict the transcript $Y_1$ from the compressed audio input.
We do not compress prior-turn text in this work. Unlike audio, compressed text did not admit a comparably effective alignment stage in our preliminary experiments, and retaining transcripts explicitly yielded a simpler and more reliable training setup.

\noindent \textbf{Stage~2: Contextual Fine-tuning.} \quad
In the second stage, we initialize the compression module from the Stage~1 checkpoint and fine-tune the model using the compressed multi-turn input $\mathcal{S}_{\text{comp}}$ in Eq.~\ref{eq:multi_turn_prompt_compressed}.
At this stage, we jointly optimize the compression module and the audio LoRA \cite{hu2022lora} parameters of the LLM, allowing the model to combine compressed context turns with the full-resolution audio of the current turn.
Following \cite{lin2025refragrethinkingragbased}, we use a curriculum over context length: training starts with no context turns and progressively increases the maximum number of available turns until the target number of context turns is reached.

\section{Experimental Setup}
\label{sec:exp_setup}
\subsection{Datasets}
Our experiments use DefinedAI\footnote{\url{https://www.defined.ai}} as the main in-domain dataset, WoW as the out-of-domain evaluation set, and LibriSpeech~960h \cite{panayotov2015librispeech} only for Stage~1 compression training. 

\noindent \textbf{DefinedAI} 
consists of scripted conversations between agents and customers.
Unless specified, all models are fine-tuned on the DefinedAI train split and selected on the dev split.
The dataset contains 17k/559/2k utterances in train/dev/test, corresponding to 40/2.25/4.5 hours and 359k/20k/42k total words.
The numbers of annotated contextual entities (measured as entity words) are 6.5k/379/727, with approximately 12.9k/700/1.4k entity words in train/dev/test.

\noindent \textbf{WoW}
is an internal dataset of real call-center conversations between agents and customers, used only for out-of-domain evaluation.
None of our models are trained on WoW.
The test set contains 1,465 utterances (1.76 hours) and 20k total words, of which 3,434 are entity words, corresponding to an entity-word ratio of 16.9\%.
We include WoW because it is more realistic than the scripted DefinedAI data and substantially richer in contextual entities.

\noindent \textbf{LibriSpeech~960h} is a large read-speech corpus that we use for additional pretraining of the audio compression module in single-turn compressed-audio experiments (Stage 1 of training in  Section~\ref{subsec:compression_train}). 
LibriSpeech is not used for Stage~2 contextual fine-tuning or for evaluation.

\noindent \textbf{Evaluation protocol.} \quad
We evaluate on the DefinedAI (in-domain) and WoW (out-of-domain) test splits; unless noted, all contextual models are trained on DefinedAI.

\subsection{Evaluation Metrics}
\label{subsec:eval_metrics}
We report standard word error rate (WER) over the full reference transcript and bias-word error rate (Bias-WER) over annotated contextual-entity tokens, such as person names, locations, and product names. Bias-WER uses the same alignment and edit-distance procedure as standard WER, but restricts scoring to the annotated entity spans.

\subsection{Implementation Details}
We use \textsc{Phi-4-Multimodal} as the backbone in all experiments. Single-turn and raw-context ASR models are fine-tuned on DefinedAI for 4 epochs using AdamW, batch size 16, learning rate $4\times10^{-5}$, weight decay $0.01$, maximum gradient norm $1.0$, and linear scheduler with 50 warmup steps. For raw-context ASR, the number of prior turns is sampled uniformly from 0 to 10 during training, teaching the model to use a variable number of context turns during fine-tuning.

For Abstract Compression, Stage~1 starts from the best DefinedAI single-turn checkpoint with the cross-attention compression module attached. The backbone is frozen, only the compression module is optimized, and the learning rate is increased to $10^{-3}$. Stage~2 initializes from the best Stage~1 checkpoint and jointly fine-tunes the compression module and audio LoRA parameters, using the same optimization settings as standard ASR fine-tuning. Unlike raw-context training, Stage~2 uses a curriculum over context length: training starts with zero prior turns and increases the maximum available context by one every 10\% of training until reaching 10 turns.

During decoding, we use a fixed number of prior turns for each evaluation setting.
This provides a controlled comparison across models and makes the effect of context length easier to interpret.
Prior-turn transcripts are provided as ground-truth text during both training and inference, and the model predicts only the current-turn transcript. Unless otherwise noted, checkpoints are selected on the DefinedAI dev split.

\section{Experimental Results}
\label{sec:results}
We organize the results around three questions. First, can the model benefit from conversational context at all? Second, can Abstract Compression retain those gains under a much smaller prior-turn audio budget? Third, what factors govern the quality-efficiency tradeoff?
\begin{table}[t]
\caption{ASR performance on the DefinedAI (in-domain) and WoW (out-of-domain) test sets. All systems use \textsc{Phi-4-Multimodal} (\textsc{Phi-4-MM}). Open-source \textsc{Phi-4-MM} denotes the original non-fine-tuned model, while the remaining configurations are fine-tuned variants. N$_{\text{train}}$ and N$_{\text{decode}}$ denote the number of context turns used for training and decoding.}
\centering
\small
\setlength{\tabcolsep}{2.5pt}
\begin{tabular}{lcccc}
\toprule
\textbf{Configuration} & N$_{\text{train}}$ & N$_{\text{decode}}$ & \textbf{WER} & \textbf{Bias-WER} \\
\midrule
\multicolumn{5}{l}{\textbf{DefinedAI Test Set} (In-domain)} \\
\midrule
Open-source \textsc{Phi-4-MM} & 0 & 0 & 13.4 & 21.0 \\
\cmidrule(lr){1-5}
\multirow{2}{*}{Single-turn} & \multirow{2}{*}{0} & 0 & 7.6 & 13.5 \\
& & 5 & 9.1 & 15.1 \\
\cmidrule(lr){1-5}
\multirow{2}{*}{Multi-turn, Raw}  & \multirow{2}{*}{$\mathcal{R}_{[0,10]}$} & 5 & 7.5 & 13.3 \\
& & 10 & \textbf{7.5} & \textbf{13.1} \\
\rowcolor[gray]{0.95} Multi-turn, Compressed & $0\!\to\!10$ & 10 & 8.0 & 13.3 \\
\midrule
\midrule
\multicolumn{5}{l}{\textbf{WoW Test Set} (Out-of-domain)} \\
\midrule
Open-source \textsc{Phi-4-MM} & 0 & 0 & 20.4 & 30.0 \\
\cmidrule(lr){1-5}
\multirow{2}{*}{Single-turn} & \multirow{2}{*}{0} & 0 & 13.4 & 25.6 \\
& & 10 & 14.5 & 27.2 \\
\cmidrule(lr){1-5}
\multirow{2}{*}{Multi-turn, Raw} & \multirow{2}{*} {$\mathcal{R}_{[1,10]}$} & 5 & 12.9 & 24.2 \\
& & 10 & \textbf{12.7} & \textbf{23.3} \\
\rowcolor[gray]{0.95} Multi-turn, Compressed & $0\!\to\!10$ & 10 & 13.2 & 24.5 \\
\bottomrule
\end{tabular}
\label{tab:context_results}
\end{table}
\subsection{Raw Context}
\label{subsec:raw_history_results}
Table~\ref{tab:context_results} compares single-turn ASR, inference-time context prompting, and raw multi-turn fine-tuning. Fine-tuning \textsc{Phi-4-Multimodal} for single-turn ASR substantially improves over the open-source model, establishing the single-turn row as the main baseline. Simply decoding this model with context, however, degrades performance, showing that inference-time prompting alone is insufficient for reliable cross-turn use.
After supervised multi-turn fine-tuning, raw context improves over the single-turn baseline on both datasets, with larger gains on Bias-WER than overall WER. On DefinedAI, 10 context turns reduce Bias-WER from 13.5\% to 13.1\%; on the more entity-dense WoW set, WER improves from 13.4\% to 12.7\% and Bias-WER from 25.6\% to 23.3\%. This supports our hypothesis that conversational context is most useful for recovering contextual entities.

A context-modality ablation on WoW (Table~\ref{tab:context_results_ablation}) shows that raw audio-only context performs similarly to audio+text context and significantly better than text-only context, suggesting that full-resolution audio already contains most contextual cues. This supports compressing prior-turn audio, rather than relying on standalone transcript conditioning, to preserve contextual information. We nevertheless retain explicit transcripts in Abstract Compression because text adds little prompt overhead and preserves the standard chat-template structure of \textsc{Phi-4-Multimodal}, ensuring stable instruction-following.

\begin{table}[t]
\caption{Context-modality ablation on WoW. N$_{\text{decode}}$ denotes the number of context turns used at decoding.}
\centering
\small
\setlength{\tabcolsep}{2.5pt}
\begin{tabular}{lccc}
\toprule
\textbf{Context Modality} & N$_{\text{decode}}$ & \textbf{WER} & \textbf{Bias-WER} \\
\midrule
\multirow{2}{*}{Audio \& Text} & 5 & 12.9 & 24.2 \\
& 10 & 12.7 & 23.3 \\
\midrule
\multirow{2}{*}{Audio only} & 5 & 13.0 & 23.8 \\
& 10 & 12.8 & 23.2 \\
\midrule
\multirow{2}{*}{Text only} & 5 & 13.5 & 28.9 \\
& 10 & 13.4 & 28.5 \\
\bottomrule
\end{tabular}
\label{tab:context_results_ablation}
\end{table}

\subsection{Abstract Compression}
\label{subsec:compressed_results}
\noindent \textbf{Stage 1.} \quad
Table~\ref{tab:stage1_alignment} evaluates whether compressed audio embeddings retain enough information for single-turn ASR. Increasing $K$ improves WER, and adding LibriSpeech~960h to DefinedAI further improves the $K=16$ setting. Although compressed audio remains worse than full-resolution audio, these results show that the latent tokens preserve useful acoustic information for downstream contextual modeling.

\begin{table}[t]
\caption{Stage~1 ASR from compressed audio (without context). $K$ denotes the number of latent audio tokens.}
\centering
\small
\setlength{\tabcolsep}{4.4pt}
\begin{tabular}{lccc}
\toprule
\textbf{Compression} & \textbf{Training} & \textbf{DefinedAI} & \textbf{WoW}\\
\textbf{Setting} & \textbf{Data} & \textbf{WER} & \textbf{WER} \\
\midrule
Full audio & DefinedAI (40h) & 7.6 & 13.4 \\
\midrule
$K=4$  & DefinedAI (40h) & 51.2 & 72.3 \\
$K=8$  & DefinedAI (40h) & 46.3 & 66.4 \\
$K=16$ & DefinedAI (40h) & 37.1 & 54.2 \\
\midrule
\multirow{2}{*}{$K=16$} & LibriSpeech (960h) & \multirow{2}{*}{\textbf{21.2}} & \multirow{2}{*}{\textbf{29.1}} \\
 & + DefinedAI (40h) & & \\
\bottomrule
\end{tabular}
\label{tab:stage1_alignment}
\end{table}
\noindent \textbf{Stage 2.} \quad
Table~\ref{tab:context_results} shows that compressed multi-turn conditioning improves Bias-WER over the single-turn baseline while remaining behind raw-context conditioning. On DefinedAI, Abstract Compression reaches 13.3\% Bias-WER, compared with 13.5\% for single-turn and 13.1\% for raw context. On WoW, it reaches 24.5\% Bias-WER, improving over the 25.6\% single-turn baseline while trailing the 23.3\% raw-context result. Thus, compressed context recovers a meaningful fraction of the entity-recognition benefit of raw context under a smaller prior-turn audio budget.


\subsection{Compression Rate Analysis}
\label{sec:comp_analysis}
\begin{figure}[t]
    \centering
    \includegraphics[width=\columnwidth]{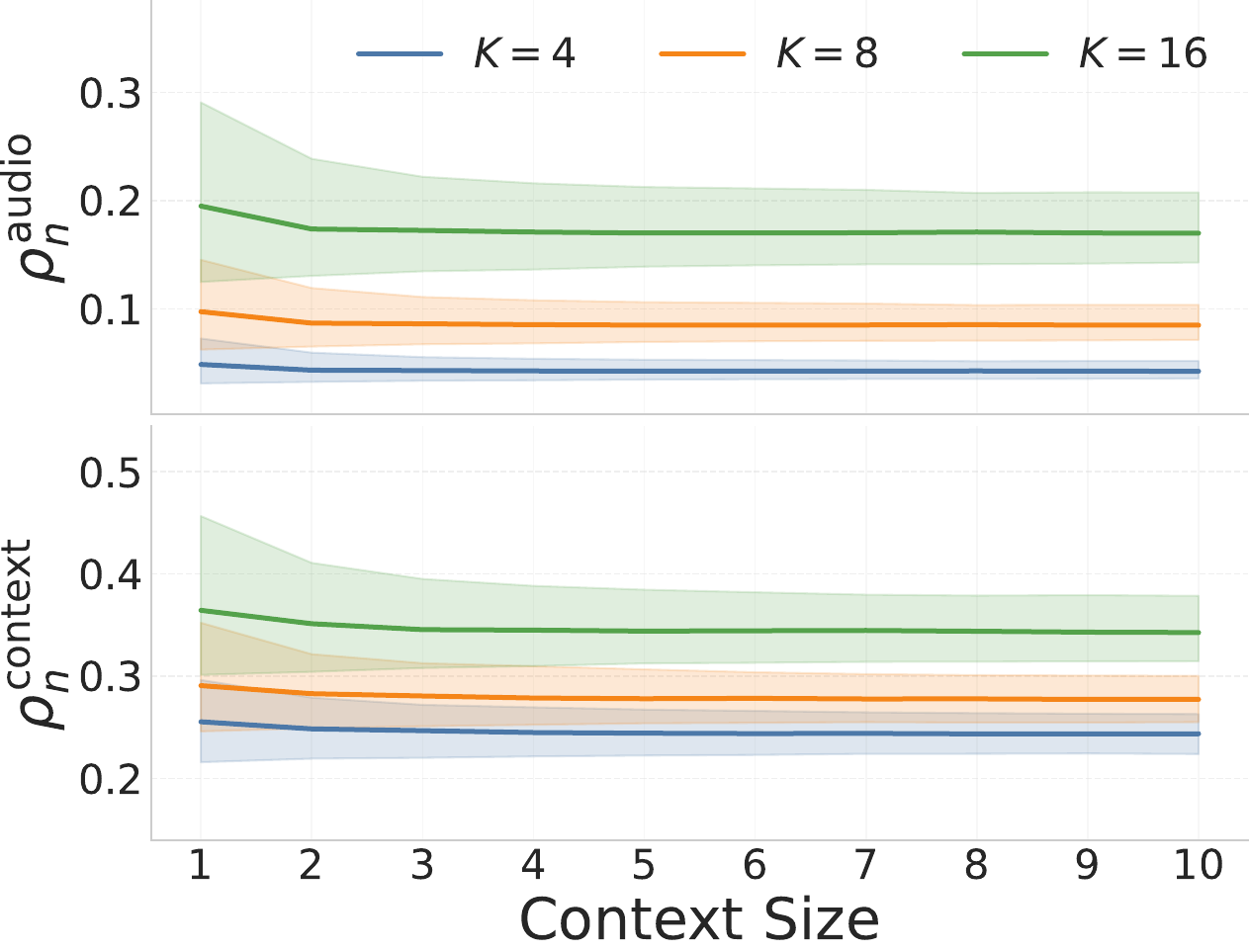}
    \caption{Compression rates as a function of context size on DefinedAI test set.
Top: audio compression rate $\rho_n^{\text{audio}}$ (see Eq.~\ref{eq:rho_aud}).
Bottom: overall context compression rate $\rho_n^{\text{context}}$ (see Eq.~\ref{eq:rho_hist}).
Lines show the median across examples and shaded regions denote interquartile range.}
    \label{fig:compression_rate_boxplot}
\end{figure}

To evaluate efficiency of Abstract Compression, we measure how much of the context input is retained after compression.
For a sample with $n$ prior turns, we first define the audio-only compression rate as
\begin{equation}
\label{eq:rho_aud}
\rho_n^{\text{audio}} =
\frac{nK}{\sum_{i=1}^{n} |\mathbf{X}_i|},
\end{equation}
where $|\mathbf{X}_i|$ is the number of raw audio embeddings in context turn $i$, and $K$ is the fixed number of latent audio embeddings per turn. This captures how aggressively the prior-turn audio stream is compressed.
Second, we define the overall context compression rate as
\begin{equation}
\label{eq:rho_hist}
\rho_n^{\text{context}} =
\frac{\sum_{i=1}^{n} \left(K + |Y_i|\right)}
     {\sum_{i=1}^{n} \left(|\mathbf{X}_i| + |Y_i|\right)},
\end{equation}
where $|Y_i|$ is the number of transcript tokens in turn $i$.
This metric reflects the retained fraction of the full context prompt, since our method compresses context audio only, while keeping prior-turn transcripts explicit. Smaller values of either ratio indicate stronger compression.
The distinction between the two metrics is important.
$\rho_n^{\text{audio}}$ measures how aggressively the audio is compressed, whereas $\rho_n^{\text{context}}$ captures the realized reduction in the full conversational context seen by the model at inference time.
Because prior-turn transcripts are not compressed, $\rho_n^{\text{context}}$ is naturally larger than $\rho_n^{\text{audio}}$.

Figure~\ref{fig:compression_rate_boxplot} shows two trends on the DefinedAI test set. First, both $\rho_n^{\text{audio}}$ and $\rho_n^{\text{context}}$ decrease and then stabilize as more context turns are included, since each additional turn contributes exactly $K$ latent audio tokens regardless of its original duration. Second, the variance of both ratios narrows with more turns, as utterance-level length variation averages out across the context. Thus, Abstract Compression not only reduces the average context token budget, but also makes the effective context footprint more predictable across examples.

\subsection{Ablation Studies}
\label{sec:ablation_studies}
We next analyze which aspects of the compression setup affect performance the most. All ablations are reported on the WoW test set in Table~\ref{tab:stage2_ablation}.

\noindent \textbf{Effect of bottleneck size.} \quad
The number of latent tokens $K$ directly controls the information capacity of the bottleneck.
The trend is consistent across both Stage~1 and Stage~2.
In Stage~1 (Table~\ref{tab:stage1_alignment}), increasing $K$ improves single-turn transcription from compressed audio alone. In Stage~2 (Table~\ref{tab:stage2_ablation}), larger $K$ also makes compressed context more useful. With $K=4$, Bias-WER improves only modestly, from 26.6\% without context to 25.2\% with context. With $K=8$, the best Bias-WER is 25.5\%.
The strongest results are obtained with $K=16$, where Bias-WER drops from 26.5\% without context to 24.5\% with 5 or 10 context turns.
This suggests that very aggressive compression removes information needed for cross-turn disambiguation, whereas a moderate bottleneck can preserve a useful portion of the contextual signal.

\noindent \textbf{Effect of Stage~1 data.} \quad
The Stage~1 results in Table~\ref{tab:stage1_alignment} show that a larger Stage~1 training set clearly helps in the single-turn alignment setting: for $K=16$, pretraining on LibriSpeech~960h + DefinedAI substantially improves Stage~1 WER relative to DefinedAI-only training.
However, Table~\ref{tab:stage2_ablation} shows that this gain does not translate into better contextual ASR after Stage~2 fine-tuning.
For $K=16$, initializing Stage~2 with LibriSpeech~960h + DefinedAI does not outperform the DefinedAI-only initialization.
This suggests that better single-turn compressed-audio reconstruction is not sufficient by itself to guarantee better downstream contextual use.
One possible explanation is that Stage~2 depends not only on whether the latent tokens preserve acoustic content, but also on whether that content is aligned with the conversational cues that matter for entity recovery in the downstream domain.
In our setting, in-domain Stage~1 training may therefore provide a better initialization for this task-specific alignment, even if it is weaker as a general-purpose audio compressor.

\begin{table}[t]
\caption{Stage~2 compression ablations on WoW.}
\centering
\small
\setlength{\tabcolsep}{3.8pt}
\begin{tabular}{lcccc}
\toprule
\textbf{Latent} & \textbf{Stage~1} & \multirow{2}{*}{\textbf{N}$_{\text{decode}}$} & \multirow{2}{*}{\textbf{WER}}& \textbf{Bias-}\\
\textbf{Tokens ($K$)} & \textbf{Train Data} & & & \textbf{WER} \\
\midrule
\multirow{4}{*}{4}  & \multirow{4}{*}{DefinedAI (40h)} & 0 & 13.5 & 26.6 \\
& & 1 & \textbf{13.2} & 26.0 \\
& & 5 & 13.3 & 25.2 \\
& & 10 & 13.4 & 25.3 \\
\midrule
\multirow{4}{*}{8}  & \multirow{4}{*}{DefinedAI (40h)} & 0 & 13.7 & 27.0 \\
& & 1 & 13.5 & 26.6 \\
& & 5 & 13.4 & 25.5 \\
& & 10 & 13.5 & 25.6 \\
\midrule
\multirow{4}{*}{16}  & \multirow{4}{*}{DefinedAI (40h)} & 0 & 13.5 & 26.5 \\
& & 1 & 13.3 & 25.3 \\
& & 5 & 13.3 & \textbf{24.5} \\
& & 10 & \textbf{13.2} & \textbf{24.5} \\
\midrule
\multirow{4}{*}{16}  & & 0 & 13.4 & 26.3 \\
& LibriSpeech (960h) & 1 & 13.3 & 25.1 \\
& + DefinedAI (40h) & 5 & 13.6 & 25.1 \\
& & 10 & 13.4 & 25.1 \\
\bottomrule
\end{tabular}
\label{tab:stage2_ablation}
\vspace{-0.5cm}
\end{table}
\noindent \textbf{Effect of context length.} \quad
Table~\ref{tab:stage2_ablation} shows how the compressed model behaves as context turns increase at inference.
In nearly all settings, adding context improves over the no-context condition, confirming that the model uses the compressed representations.
Most gains arise from the first few turns and are largely saturated with 5 turns.
For instance, with $K=16$ and DefinedAI Stage~1 training, Bias-WER drops from 26.5\% with 0 context turns (single-turn) to 25.3\% with 1 turn and 24.5\% with 5 turns, with no additional improvement with 10 turns.
This pattern is favorable from an efficiency standpoint, since it suggests that a moderate amount of compressed context captures most of the contextual benefit.

\noindent \textbf{System-Level Efficiency Implications.} \quad
We benchmark the LLM forward pass after input construction to isolate the cost induced by the input sequence on a single NVIDIA H100 80 GB GPU. Table~\ref{tab:system_efficiency} compares single-turn decoding with raw and compressed 10-turn decoding, showing raw-context substantially increases both forward time and peak GPU memory relative to the single-turn baseline, while Abstract Compression reduces this overhead significantly.

\begin{table}[H]
\caption{Abstract Compression efficiency on the test set. Avg. time is wall-clock LLM-forward time per decoded utterance, reported in seconds; peak memory is reported in GB, and $K$ denotes the number of latent audio tokens per context turn.}
\centering
\small
\setlength{\tabcolsep}{3.2pt}
\begin{tabular}{lcccc}
\toprule
\textbf{Configuration} & 
\textbf{$K$} & 
\textbf{N}$_{\text{decode}}$ & 
\textbf{Avg. Time} & 
\textbf{Peak Mem.} \\
\midrule
Single-turn & -- & 0 & \textbf{0.440} & \textbf{11.07} \\
Multi-turn, Raw & -- & 10 & 1.525 & 28.98 \\
\rowcolor[gray]{0.95}
Multi-turn, Compressed & 4 & 10 & \textbf{0.644} & \textbf{17.13} \\
\rowcolor[gray]{0.95}
Multi-turn, Compressed & 8 & 10 & 0.732 & 17.65 \\
\rowcolor[gray]{0.95}
Multi-turn, Compressed & 16 & 10 & 0.835 & 18.05 \\
\bottomrule
\end{tabular}
\label{tab:system_efficiency}
\end{table}

\section{Illustrative Example}
\label{sec:qualitative}
Figure~\ref{fig:qualitative_example} demonstrates how acoustic context aids recognition: when the entity ``Buick Encore'' appears in the history and is spoken rapidly in the target turn, the single-turn model predicts the phonetically similar phrase ``VIEWER CONCOURSE,'' whereas both raw and compressed context recover the correct entity. This suggests that compressed context can preserve acoustic cues useful for recovering the entity.

\begin{figure}[H]
\centering
\begin{tcolorbox}[
    enhanced,
    width=\columnwidth,
    colback=white,
    colframe=black!25,
    colbacktitle=black!12,
    coltitle=black,
    title={Compressed context recovers the entity},
    fonttitle=\bfseries\footnotesize,
    boxrule=0.45pt,
    arc=2pt,
    left=5pt,
    right=5pt,
    top=4pt,
    bottom=4pt,
    toptitle=3pt,
    bottomtitle=3pt,
    titlerule=0pt
]
\fontsize{7.2}{8.4}\selectfont
\setlength{\tabcolsep}{0pt}
\renewcommand{\arraystretch}{1.18}

\begin{tabularx}{\linewidth}{
@{}>{\bfseries\raggedright\arraybackslash}p{0.20\linewidth}
   >{\raggedright\arraybackslash}X@{}
}
History &
HOW MUCH FOR THE TWENTY \textbf{BUICK ENCORE} \\

Target &
THIRTY ONE FOR THE TWENTY \textbf{BUICK ENCORE} \\

Single-turn &
THIRTY ONE FOR THE TWENTY \textcolor{red}{\textbf{VIEWER CONCOURSE}} \\

Raw context &
THIRTY ONE FOR THE TWENTY \textcolor{blue}{\textbf{BUICK ENCORE}} \\

Compressed &
THIRTY ONE FOR THE TWENTY \textcolor{blue}{\textbf{BUICK ENCORE}}
\end{tabularx}
\end{tcolorbox}

\vspace{-1.5mm}
\caption{Qualitative entity-recovery example.}
\label{fig:qualitative_example}
\end{figure}

\section{Conclusion}
We show that conversational multimodal context can improve LLM-based ASR, but only after supervised multi-turn training, with the largest gains accruing on contextual entities. To reduce the cost of raw-context conditioning, we introduce Abstract Compression, which compresses prior-turn audio into a fixed latent budget and recovers part of the benefit of full raw context. Overall, our results suggest that compact representations of conversational audio context offer a practical quality-efficiency tradeoff for context-aware ASR.





\section*{Acknowledgements}
This work was supported by Idiap Research Institute and Uniphore collaboration project. Part of this work was also supported by EU Horizon 2020 project ELOQUENCE (grant number 101070558).

The authors used ChatGPT (OpenAI) and Claude (Anthropic) for language editing and improving the clarity of the manuscript. All experiments, results, and scientific claims are the authors' own.

\bibliographystyle{IEEEtran}
\bibliography{arr2026/custom}

\end{document}